\documentclass[journal,letterpaper]{IEEEtran}
\usepackage[T1]{fontenc}
\usepackage{amsmath,amssymb,graphicx,booktabs,array}
\usepackage{cite}
\usepackage[hidelinks]{hyperref}
\usepackage{url}
\hypersetup{pdftitle={Episode-Calibrated Action Prefixes from Robot\textendash{}Object Graph Forecasts},
  pdfauthor={Aobo Li, Ying Liu, Yunjian Zhang, Yilun Wu, Mu-Jiang-Shan Wang}}
\begin{document}
\bstctlcite{IEEEexample:BSTcontrol}
\title{Episode-Calibrated Action Prefixes from Robot--Object Graph Forecasts}
\author{Aobo~Li, Ying~Liu, Yunjian~Zhang, Yilun~Wu, and Mu-Jiang-Shan~Wang,~\IEEEmembership{Member,~IEEE}%
\thanks{\raggedright Aobo Li, Yunjian Zhang, and Yilun Wu are with Shenzhen Kaihong Digital Industry Development Co., Ltd., Shenzhen, China. A. Li (\href{mailto:ali09@qub.ac.uk}{\mbox{ali09@qub.ac.uk}}, ORCID \href{https://orcid.org/0009-0001-6868-9495}{\mbox{0009-0001-6868-9495}}). Y. Zhang (\href{mailto:yunjzhang@xmu.edu.cn}{\mbox{yunjzhang@xmu.edu.cn}}, ORCID \href{https://orcid.org/0000-0002-0303-3671}{\mbox{0000-0002-0303-3671}}). Y. Wu (\href{mailto:wuyilun310@gmail.com}{\mbox{wuyilun310@gmail.com}}, ORCID \href{https://orcid.org/0000-0001-5169-4864}{\mbox{0000-0001-5169-4864}}).}%
\thanks{\raggedright Ying Liu is with Southern University of Science and Technology, Shenzhen, China (\href{mailto:Liuy63@sustech.edu.cn}{\mbox{Liuy63@sustech.edu.cn}}, ORCID \href{https://orcid.org/0000-0001-6016-3184}{\mbox{0000-0001-6016-3184}}).}%
\thanks{\raggedright Mu-Jiang-Shan Wang (corresponding author) is with Shenzhen Kaihong Digital Industry Development Co., Ltd., Shenzhen, China, and with Harbin Institute of Technology, China (\href{mailto:wangmujiangshan-qy@kaihong.com}{\mbox{wangmujiangshan-qy@kaihong.com}} and \href{mailto:wmjsdr0@gmail.com}{\mbox{wmjsdr0@gmail.com}}, ORCID \href{https://orcid.org/0000-0003-0950-4558}{\mbox{0000-0003-0950-4558}}).}}
\maketitle
\begin{abstract}
An action-chunk controller must decide how long to commit before observing the scene again. We calibrate a simultaneous upper bound on the largest robot--object landmark forecast error and release the longest prefix below a task tolerance. Treating each complete trajectory as the calibration unit permits selection at prespecified decision windows under episode exchangeability. On 206 Push-T demonstrations, episode calibration at 32 pixels released 4.03 of eight steps on average and exceeded the tolerance in four of 60 test episodes. Pooling windows released 6.92 steps and exceeded it in 29 episodes. On 206 physical pushes of a rectangle, a horizon-only scale admitted longer prefixes than the graph-dependent scale. In simulator rollouts, recalibrating a retrieval policy on separate resets reduced violations from 15 to two in 60 episodes while raising calls per 100 actions from 29.0 to 55.1. A post-observation trigger gave one violation at 50.9 calls. Frozen image-conditioned policies further separated geometric error from task success. On 24 matched resets, fixed-eight and calibrated Diffusion Policy completed 2 and 3 episodes, with 21 and 0 episodes exceeding the authorized tolerance. Subsampling the physical calibration pool to 20 episodes gave a central 80\% range of 4.02--5.67 released steps at 2 mm. The rank guarantee applies to the stated episode distribution. Each new execution rule requires a closed-loop audit.
\end{abstract}

\section{Introduction}
\IEEEPARstart{A}{ pusher} approaching the stem of a T-shaped object can move several pixels while the object stays still. Near an edge of the crossbar, a similar movement can rotate the object enough to displace a distant landmark. A fixed action horizon treats the two situations alike. A controller with a forecast of landmark motion can instead ask how many commands it can issue before its geometric error budget is exhausted.

PI-VLA monitors disagreement between action representations and state predictions to vary its execution horizon \cite{pi}. Real-time chunking \cite{realtime} and masked action chunking \cite{remac} address the interaction between the next command and work already in progress. FlashVLA streams the decoder state through asynchronous execution \cite{flash}. These mechanisms make horizon selection practical. To attach a probability statement to the selection, one must first specify the event being counted. The chance of a large error in a randomly selected window differs materially from the chance that any committed command in a whole manipulation episode exceeds its tolerance.

We examine the second event. A point forecast supplies one future position for every landmark and horizon. A scale predictor estimates which forecasts are likely to have large residuals. Calibration uses one maximum per demonstration. At a decision time, the controller takes the longest prefix whose bounds fit within the specified tolerance. The statistical ingredient is a standard exchangeable-rank argument. The method contribution lies in how the graph residual, decision windows and time of commitment define one event that remains covered after a prefix has been selected.

The empirical question is how much usable prefix length remains after paying for simultaneous coverage. We compare pooled-window calibration, episode calibration with a common scale at each horizon, and episode calibration with geometry-dependent scales. The different methods use identical point forecasts and replayed commands. The release includes the split assignments and each held-out calibration and test residual so the order statistic and every reported decision can be reconstructed.

\section{Related Work}
\subsection{Forecasts available to a chunk policy}
Action chunks appear in bimanual imitation \cite{act} and diffusion policies \cite{diff}. Large policies such as OpenVLA, RT-2 and Octo change the distribution and representation of a proposed action sequence \cite{openvla,rt2,octo}. PerAct represents 3-D action targets explicitly \cite{peract}. In a replay study the logged command supplies an unambiguous candidate. Applying the same score to samples from a pretrained policy would require calibration trajectories from that policy or assumptions linking the two distributions.

Latent visual dynamics can be trained for planning \cite{pla,lewm}, while world models can also be optimized to support control through imagined returns \cite{dream}. Motus integrates world and action prediction \cite{motus}; BagelVLA interleaves its forecasts with language and action \cite{bagel}. FIRM-WM collects paired alternatives at common resets \cite{firm}, a useful way of obtaining labels for actions that are absent from a demonstration log. SIMPACT uses simulation to rank actions \cite{simpa}. All can supply candidate futures; none of these choices by itself specifies an episode-level bound on selected physical errors.

\subsection{Graph residuals and the time of detection}
Robot-WATCHDOG is the closest recent comparison \cite{watch}. It predicts object-centric dynamics or compares object trajectories against expert examples, then calibrates a failure alarm. Its labeled outcome is an anomaly in the trajectory. Our decision precedes observation of the future object state, and its target is the length of a prefix of a given chunk under a positional tolerance. One can have a nominal task execution with a large forecast error, or an accurately forecast trajectory that fails the task. A false-alarm bound for an anomaly detector and a bound on accepted forecast error answer different questions.

Interaction networks and learned physics graphs supply local motion messages \cite{inter,gns,mpnn}. MomaGraph represents instruction-relevant object states \cite{moma}, and ConceptGraphs builds open-vocabulary scene structure \cite{concept}. We use a much smaller graph with known landmark identities to isolate uncertainty from data association. In a camera-based system, matching and localization enter before calibration. HybridGNN addresses matching \cite{hybrid}, and MFE-DETR concerns cross-modal object detection \cite{mfe}. A switch of tracker changes the distribution of the measured endpoint. Persistent errors in identity have a different structure from independent missing edges, a distinction also familiar in network diagnosis under constrained faults \cite{diag}.

FSTGAT models time-varying relational dependence in a financial setting \cite{fstgat}; factorized multimodal learning separates complementary measurement channels \cite{factor}. We take a limited design cue from those studies: the scale used for uncertainty can use features different from the point predictor. Thermal model predictive control provides another instance in which sensing and a future tolerance are connected at decision time \cite{thermal}. These cross-domain examples motivate architectural choices. The episode-rank result below depends only on the sampling assumptions.

\subsection{Calibration after selection}
Uncertainty estimates from ensembles \cite{ensem} and confidence calibration for neural classifiers \cite{temp} give useful rankings, whose relationship to physical error must be measured. Conformalized quantile regression permits a learned local width \cite{cqr}; predictive inference with the jackknife+ exploits refitting across folds \cite{jack}. Conformal risk control extends ranking ideas to monotone losses \cite{crc}. KnowNo uses conformal uncertainty to decide when an embodied planner should ask for assistance \cite{knowno}. In our setting the fallback is a shorter prefix, including length zero, and the loss is an exceedance of a geometric tolerance.

Multi-horizon conformal forecasting \cite{rnn,confme}, multivariate regions \cite{multid}, and relational conformal prediction \cite{relconf} address dependencies across forecast targets. Work on dependent time series and changing distributions \cite{traj,why,adaptive,beyond} addresses assumptions that differ from exchangeability of complete episodes. We allow arbitrary dependence inside each demonstration. We pay for simultaneous validity by compressing it to its worst standardized residual. This sacrifices width whenever a single contact transition is difficult to forecast.

\section{Graph Forecast and Commitment Rule}
\subsection{Information available before commitment}
Episode $i$ consists of a pusher position $p_t$, nine tracked object landmarks $x_{jt}$, and a proposed command sequence $a_{t:t+H-1}$. A command in Push-T is a planar pusher target. We condition on the supplied chunk, including its future targets. Measured future pusher or object positions are excluded. At test time each prefix is a recommendation on this fixed chunk. Replanning after executing a different prefix would generally generate a different trajectory.

For each horizon $h\le H$, define a predicted displacement $f_{jh}$ at every landmark. Edges connect the pusher to the landmarks. In units of 100 scene pixels, let $r_j=(p_t-x_{jt})/100$, $c_j=(\bar x_t-x_{jt})/100$, $d_h=(a_{t+h-1}-p_t)/100$ and $w_j=\exp(-\|r_j\|^2/0.45^2)$. A shared nonlinear edge message and pooled graph message are
\begin{equation}
 m_{jh}=w_j(d_h-r_j),\qquad
 \bar m_h=9^{-1}\sum_{j=1}^{9}m_{jh}.
 \label{eq:mess}
\end{equation}
The shared landmark readout is linear in $r_j,c_j$, landmark and pusher velocities, the final and mean command offsets, $w_j,w_jd_h,w_jr_j$, and $\bar m_h$, with an intercept. Velocity features are adjacent-frame displacements divided by 10. The readout predicts displacement in scene pixels separately at each horizon by ridge regression. It is a one-round graph predictor with a fixed radial message kernel and learned output weights. Relative positions remove translation of the scene; summation and shared weights commute with landmark reindexing. No image encoder is used.

\begin{figure}[!t]\centering\includegraphics[width=.96\columnwidth]{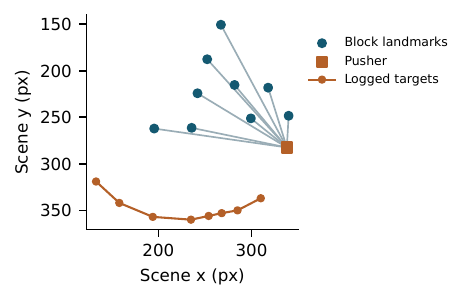}
\caption{An actual Push-T decision window from the midpoint of the first held-out episode in the primary permutation. The graph joins the pusher to nine landmarks. The orange path contains the eight logged command targets available before commitment. Coordinates are in the original scene frame.}
\label{fig:graph}\end{figure}

At a window $t$ the scalar residual is the largest error across the landmarks,
\begin{equation}
 e_{ith}=\max_j \|x_{j,t+h}-x_{jt}-f_{jh}\|_2.
 \label{eq:e}
\end{equation}
An orientation error can move one end of the block while the centroid changes little. The maximum makes such a rotation visible. A single corrupted landmark can also dominate it. A quantile or trimmed maximum would correspond to a different observation contract.

\subsection{Fitting a scale without using calibration labels}
A second set of training episodes supplies labels $\log(1+e_{ith})$ for a ridge regression at each horizon. Predictors are an intercept, final and mean command-offset norms, pusher speed, maximum and mean $w_j$, largest landmark speed and the product of maximum $w_j$ with the final offset norm. The fitted log value is clipped to $[0,6]$ and transformed via $\exp(\cdot)-1$, with a one-pixel floor. Denote the result $s_{ith}>0$. Both ridge penalties are 1, applied to the unnormalized sum of squared errors including the intercept. All fits are frozen before calibration episodes are examined. A flat comparator replaces $s_{ith}$ by the horizon-wise median residual on the same scale-fitting set, also floored at one pixel.

Fix a predetermined set of anchors $\mathcal T_i$ for each episode. Its size may vary, provided the anchor-generation rule is the same for calibration and test. For $m$ calibration episodes compute
\begin{align}
 S_i &= \max_{t\in\mathcal T_i,\ 1\le h\le H} e_{ith}/s_{ith},\\
 k &=\lceil(m+1)(1-\alpha)\rceil,\qquad q=S_{(k)}.
 \label{eq:rank}
\end{align}
Set $q=+\infty$ when $k>m$. At a new anchor precompute $B_{th}=qs_{th}$. Given tolerances $\tau_{th}$, release
\begin{equation}
 L_t=\max\bigl(\{0\}\cup
 \{\ell\le H: B_{th}\le \tau_{th}\ \forall\,1\le h\le\ell\}\bigr).
 \label{eq:prefix}
\end{equation}
For a common tolerance this is a comparison against the running maximum of the bounds. A poor forecast for the first step correctly blocks a seemingly easy later step. The comparison takes $O(H)$ work after the forecasts and scales have been evaluated.

\subsection{Why the selection retains coverage}
\emph{Proposition 1.} Conditional on the fitted predictor and scale, assume the $m$ calibration episodes and one further episode are exchangeable and scored by the same rule. With probability at least $1-\alpha$, for every anchor of the new episode and every $h\le L_t$ the realized error satisfies $e_{th}\le\tau_{th}$.
\begin{IEEEproof}
Assigning independent random tie breakers gives the test score a uniform rank among the $m+1$ episode scores. When its rank is at most $k$, its score is at most $q$. Ignoring the auxiliary tie breakers can only enlarge this event. Hence all its errors simultaneously satisfy $e_{th}\le qs_{th}$ with probability at least $k/(m+1)$ if $k\le m$, and with probability one for the infinite threshold. Equation~\eqref{eq:prefix} places every released horizon inside this envelope. Taking a subset of the covered coordinates preserves simultaneous coverage.
\end{IEEEproof}

This proof allows tolerance selection from currently observed geometry and the proposed commands. Its guarantee is marginal over a fresh episode and the calibration sample. It gives no conditional $90\%$ guarantee on the relatively rare episodes that contain severe occlusion. Nor does it extend automatically to a sequence of new policy actions generated after early stopping. Logged trajectories provide future labels for the recorded branch only.

A small calculation explains why pooled-window calibration can be misleading. Even with independent windows and exactly $90\%$ coverage per window, ten windows give probability $1-0.9^{10}\simeq0.65$ of at least one miss. Adjacent robot windows are generally dependent, so this numerical value is illustrative. It identifies which risk changes when the unit of calibration changes.

\begin{figure*}[!t]\centering\includegraphics[width=.94\textwidth]{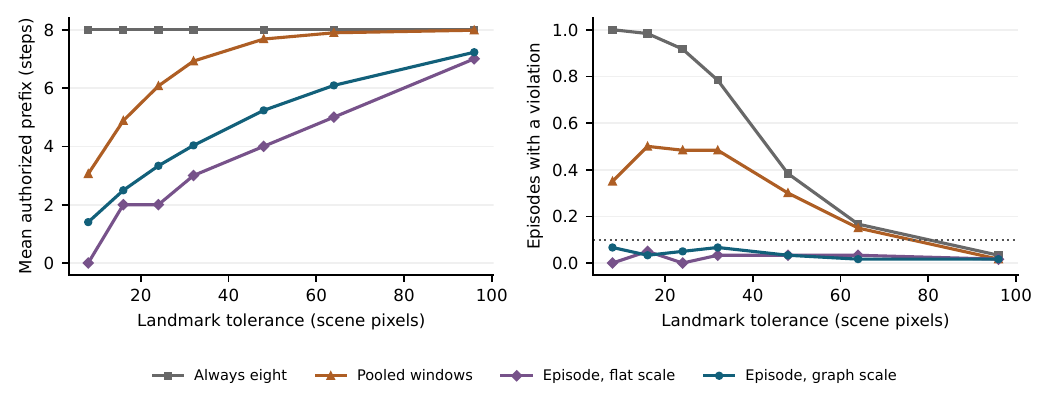}
\caption{Operating curves on the primary split of 60 test episodes. Left, mean number of logged commands within the rule's budget. Right, fraction of episodes containing at least one accepted exceedance. The dotted line marks the nominal 0.10 level. Both the authorized set and the exceedance criterion change with tolerance.}
\label{fig:trade}\end{figure*}

\section{Public Replay Study}
\subsection{Splits and measurements}
We used the original low-dimensional Push-T archive released with Diffusion Policy \cite{diff}. It contains 25,650 frames from 206 teleoperated planar simulation episodes. We used the nine recorded block landmarks, pusher center and action targets. Pixels refer to the original scene-coordinate system. They cannot be read as distances on a physical robot without an additional calibration. The released script verifies the archive checksum.

We shuffled complete episode IDs with seed 1024 and allocated 66 to point fitting, 20 to scale fitting, 60 to calibration and 60 to testing. The first anchor is local frame 1 so that a previous observation exists. Later anchors are eight frames apart, with a full eight-step future required. This yielded 1,010 fitting windows, 293 scale windows, 877 calibration windows and 884 test windows. Adjacent windows within an episode remain grouped. For $\alpha=0.10$, the calibration threshold is the 55th smallest of 60 maxima.

We compared four rules. Always-eight commits the full logged sequence. Pooled windows computes one maximum across horizons per calibration window and ranks the resulting 877 scores. Flat- and graph-scale episode methods both rank one maximum per episode. We evaluated seven tolerances from 8 to 96 pixels. The reported operating point is 32 pixels; the complete curve is shown so that its selection does not conceal a loss of coverage elsewhere. Prefix and abstention averages first weight windows within an episode equally and then weight episodes equally. A violating episode contains at least one accepted horizon with $e_{ith}>\tau$. Wilson intervals describe uncertainty in the test-episode proportion at a fixed calibration fit. We used 11 more permutations, seeds 1025--1035, to measure split sensitivity.

\section{Results}
\begin{table}[!t]\centering\caption{Primary split at 32 scene pixels. Simultaneous coverage counts full envelopes, including horizons withheld from execution.}\label{tab:main}
\footnotesize\begin{tabular}{@{}lrrr@{}}\toprule
Rule & Mean prefix & Violations / 60 & Covered / 60\\\midrule
Always eight &8.00&47&---\\
Pooled windows &6.92&29&15\\
Episode, flat scale &3.00&2&55\\
Episode, graph scale &4.03&4&51\\\bottomrule\end{tabular}\end{table}

Window pooling produced a standardized threshold of 2.39, versus 6.32 for the episode maximum with identical graph scales. The former admitted almost seven steps on average. Nearly half of the test episodes then had at least one positional violation. The graph-scale episode method released just over four steps, with four violating episodes (6.7\%, 95\% Wilson interval 2.6--15.9\%). The corresponding flat rule released three and had two violating episodes (3.3\%, interval 0.9--11.4\%). The additional accepted step has a measurable cost on this split; the intervals are too wide to resolve a small difference in risk between the episode methods.

Only 51 of 60 graph-scale test episodes lay wholly inside their predicted envelopes. That 85\% realized coverage on one split is compatible with a marginal finite-sample guarantee. Of the nine episodes with an envelope miss, four had a miss large enough on a released coordinate to cross 32 pixels. The distinction matters for interpreting a monitor. A conservative rejection can remove an outlier from the commitment metric without repairing its forecast interval.

\begin{table}[!t]\centering\caption{Episode graph scale at selected tolerances, primary split. Abstention is the episode-weighted proportion of zero-length prefixes.}\label{tab:tol}
\footnotesize\begin{tabular}{@{}rrrr@{}}\toprule Tolerance (px)&Mean prefix&Violations / 60&Abstention\\\midrule
8&1.40&4&14.8\%\\16&2.49&2&0.2\%\\
24&3.33&3&0.0\%\\32&4.03&4&0.0\%\\
48&5.23&2&0.0\%\\64&6.09&1&0.0\%\\96&7.23&1&0.0\%\\\bottomrule\end{tabular}\end{table}

Mean authorized length increases monotonically as the geometric tolerance is relaxed. The observed violation count need not be monotone because previously withheld horizons become eligible while the event threshold rises. At 8 pixels abstention consumed 14.8\% of windows on an episode-weighted basis. At 32 pixels neither episode method needed to return an empty prefix on this split. A deployed policy would still require an explicit response for $L_t=0$, such as acquiring a new observation.

\begin{table}[!t]\centering\caption{Means across 12 complete episode permutations at 32 pixels. Splits overlap and these are descriptive sensitivity measures.}\label{tab:split}
\footnotesize\begin{tabular}{@{}lrrr@{}}\toprule Rule&Prefix&Violation rate&Envelope coverage\\\midrule
Always eight&8.00&73.6\%&---\\
Pooled windows&6.81&42.4\%&23.9\%\\
Episode, flat scale&3.00&1.3\%&89.9\%\\
Episode, graph scale&3.75&1.0\%&90.4\%\\\bottomrule\end{tabular}\end{table}

Across permutations the pooled method covered an entire episode envelope only 23.9\% of the time. The graph-scale version covered 90.4\% and produced longer prefixes than the horizon-only scale on average. Reused observations couple these estimates. Treating the 12 runs as 720 independent robot episodes would overstate precision. The primary split is retained in Table~\ref{tab:main}, including its less favorable envelope coverage.

\section{Independent Replay of Physical Pushes}
\subsection{Selection and metric endpoint}
The MIT MCube planar pushing dataset records object and pusher pose under controlled straight-line pushes \cite{mcube}. Its repeatability experiment underlies a probabilistic model of pushing dynamics \cite{pushprob}. We drew one trial from each of 206 distinct contact-position and angle combinations for a rectangular object on plywood. Using the authors' list of unreliable files, we excluded flagged records, took the first remaining repetition in lexical order per numeric condition, restricted angles to $[-1.5,1.5]$ rad, and uniformly sampled conditions without replacement with seed 6174. No repetition of a selected condition appears in another split. The supplement lists all filenames and hashes.

We synchronized tip and object poses at 50 Hz by carrying forward the last time-stamped sample from each instrument on their overlapping time interval. Nine virtual probes at a $3\times3$ grid with body-frame offsets $-30,0,30$ mm move with the measured object pose. These points make angular error measurable in millimeters. They are not independent sensor measurements. The dataset contains measured tip motion and a prescribed straight push, without a logged eight-command policy chunk. At issue time we extrapolated the last two tip positions at constant velocity to supply a nominal continuation. No future tip measurement enters that proxy. This tests the forecast envelope under straight physical motion; it cannot test how a newly generated VLA chunk changes the physical outcome.

One internal coordinate unit was 0.25 mm. This sets the radial kernel scale to 11.25 mm and the scale floor to 0.25 mm. We refit the same predictor, scale and calibration rule within this corpus, retaining the 66/20/60/60 episode allocation. The data provided 9,971 synchronized frames and 1,090 decision windows. At the primary physical split (seed 2048), 315 windows belonged to the 60 test episodes. Eight steps span 160 ms.

\begin{table}[!t]\centering\caption{Physical pushes, primary metric replay. Each pair gives mean prefix and number of violating test episodes out of 60.}\label{tab:physical}
\footnotesize\begin{tabular}{@{}lrrrr@{}}\toprule
&\multicolumn{2}{c}{1 mm}&\multicolumn{2}{c}{2 mm}\\
Rule&Prefix&Violations&Prefix&Violations\\\midrule
Always eight&8.00&47&8.00&25\\
Pooled windows&4.69&24&7.51&22\\
Episode, flat scale&0.00&0&5.00&1\\
Episode, graph scale&0.00&0&4.51&5\\\bottomrule\end{tabular}\end{table}

\subsection{What transferred across apparatus}
At 2 mm the graph rule released 4.51 steps with five violating episodes (8.3\%, 95\% Wilson interval 3.6--18.1\%). The flat rule released five steps with one violation. At 1 mm both episode-level rules abstained throughout the primary test set. Their zero violations at that tolerance have no operational benefit. Pooled windows retained useful length but exceeded the budget in 24 episodes. Over 12 permutations with seeds 2048--2059, the 2-mm prefix means were 4.28 for graph scales and 4.83 for flat scales; the corresponding violation proportions averaged 3.2\% and 3.1\%. The ordering by usable length is opposite to the Push-T result. This regular straight-line experiment supplies less evidence for an advantage from a learned local scale than the T-block demonstrations.

The physical conditions come from a designed grid measured on a shared apparatus. Random partitioning of the finite set supports a held-out-condition replay calculation. It supplies weaker evidence for exchangeability of future pushes on a different surface or robot. The causal resampling rule introduces instrument-dependent lag. Timestamp synchronization and measurement latency would need to be verified on a new apparatus.

\section{Executing the Selected Prefix}
Replay fixes the action sequence even when the controller would have replanned. We therefore issued commands to a running Push-T simulator and recorded the state after each executed action. All controllers used identical evaluation seeds and terminated at the first overlap above 95\% or at 160 actions. A frozen nearest-demonstration policy indexed eight-step chunks from the 66 fit trajectories. Its search key includes pusher position in the block frame, block position and the sine and cosine of block orientation. It maps the retrieved commands through the current block pose. It sees the current pose and uses no future observation. A rigid transform of nine body-frame points, fixed from the first demonstration frame, supplies the forecasts' landmarks. The policy is purposely inexpensive so that short commitments can be compared over many complete trajectories.

Two thresholds expose a distribution change. The demonstration-based rule retains $q=6.317$ from the primary replay. For a policy-conditioned reference, we ran the frozen policy with an eight-step commitment on 48 separate simulator resets. Taking the episode maximum of the normalized errors observed through termination gave $q=12.991$. Both gates use 32 px. If a gate authorizes zero steps, a one-step policy fallback prevents deadlock; it is counted separately and its physical outcome is also audited. The 60 evaluation resets were disjoint from the 48 calibration resets. Changing the commitment rule changes the future states that are visited, so this second threshold also needs an independent audit under the adaptive controller.

\begin{table}[!t]\centering\caption{Closed-loop Push-T, 60 paired resets per controller. Calls are policy queries per 100 executed actions. A violation is a landmark forecast error above 32 px during an authorized step.}\label{tab:loop}
\footnotesize\begin{tabular}{@{}lrrrr@{}}\toprule
Commitment & Success & Peak overlap & Calls & Violations\\\midrule
One step & 5/60 & .765 & 100.0 & 1\\
Four steps & 6/60 & .822 & 25.0 & 29\\
Eight steps & 8/60 & .808 & 12.6 & 50\\
Demonstration threshold & 6/60 & .793 & 29.0 & 15\\
Policy threshold & 7/60 & .789 & 55.1 & 2\\
Reactive trigger, 32 px & 5/60 & .813 & 13.1 & 50\\
Reactive trigger, 1 px & 3/60 & .783 & 50.9 & 1\\\bottomrule\end{tabular}\end{table}

The policy-conditioned threshold authorized 1.81 actions per call, including empty prefixes in the denominator. It required 31 single-action fallbacks in 5,042 calls. Counting every executed action, including fallbacks, still gave two violating episodes. Relative to the demonstration threshold, the violation proportion changed by $-21.7$ percentage points (95\% paired bootstrap interval $[-33.3,-10.0]$ over resets, 5,000 resamples). Task success changed by $+1.7$ points, with interval $[-5.0,8.3]$. The data identify a cost in extra policy calls and an improvement in geometric error control. For two violations in 60 episodes, the one-sided 95\% exact binomial upper limit is 10.12\%, just above the nominal 10\% target. With only five to eight successful episodes per condition, they do not establish an improvement in manipulation success.

To compare with a controller that sees errors after each action, we implemented the residual-trigger part of PI-VLA's sequential decision rule \cite{pi}. It executes the next action, observes its landmark residual, and replans if that residual exceeds a threshold. The calculation uses exactly the retrieval policy and the forecast available to the prefix gate. PI-VLA's action-head disagreement and value head require a trained CMS model and are absent here. At a 32-pixel trigger the first exceedance had already happened in 50 episodes. On 20 additional development resets, we selected the threshold closest to the query rate of the policy-conditioned gate from $\{1,2,4,8,16,32\}$ pixels. The resulting 1-pixel trigger yielded only one violating test episode at 50.9 calls per 100 actions, compared with two at 55.1 for advance authorization. This reactive comparator can be attractive when every intermediate observation is available. Only the prefix rule specifies an intended commitment before those observations arrive.

Execution horizon and denoising effort are independently adjustable in flow-matching policies. Lee \emph{et al.} measured selected settings against a task-success margin on LIBERO \cite{budget}. Here we measure the worst authorized geometric error during each episode as well as success. A policy can satisfy one endpoint while missing the other.

\subsection{Two frozen policies that see the scene}
We next replaced the retrieval policy with a public SmolVLA checkpoint trained on Push-T \cite{smolvla}. The network saw a $96\times96$ RGB image, the pusher's position and the dataset's fixed task instruction. We retained its ten flow steps and used the first eight outputs of each 50-action prediction. The graph forecast still received simulator landmarks. A second visual controller used LeRobot's 175,000-step Push-T Diffusion Policy \cite{diff}, its original input and output normalization, two observed frames, and all 100 denoising steps. It has no language input. Its checkpoint and the SmolVLA training configuration are pinned in the supplement. The latter configuration specifies 1,000 training steps. As an image-interface check, we restored the measured pusher and block poses at the midpoint of each of the 60 held-out demonstrations and regenerated the RGB image. The mean absolute channel difference from the archived images was 0.077 on the 0--255 scale. This checks renderer alignment at matched poses while leaving policy quality an empirical question.

For each policy, 24 fresh fixed-eight episodes supplied a calibration order statistic. A separate set of 24 resets supplied the common initial states for the comparisons in Table~\ref{tab:visual}. Each run had at most 120 actions, compared with 300 in the upstream Diffusion Policy evaluation configuration. Stochastic proposals were indexed by reset seed and simulation step. A new query could therefore change the trajectory without changing the initial state or access to future observations. The two models used the same 24 calibration and 24 test seeds. The policy-specific thresholds were 12.61 for SmolVLA and 8.58 for Diffusion Policy.

\begin{figure}[!t]\centering\includegraphics[width=\columnwidth]{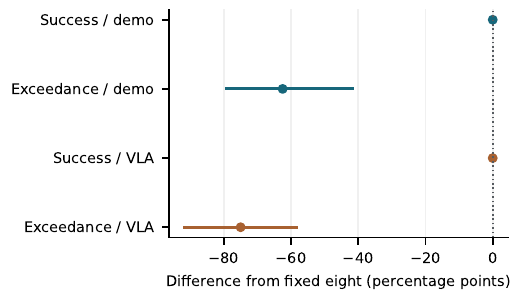}
\caption{SmolVLA paired changes relative to a fixed eight-step commitment. Intervals resample the 24 reset-level differences 5,000 times. A degenerate success interval describes the observed pairs and gives no population guarantee.}\label{fig:smol}\end{figure}

\begin{figure*}[!t]
\centering\includegraphics[width=0.90\textwidth]{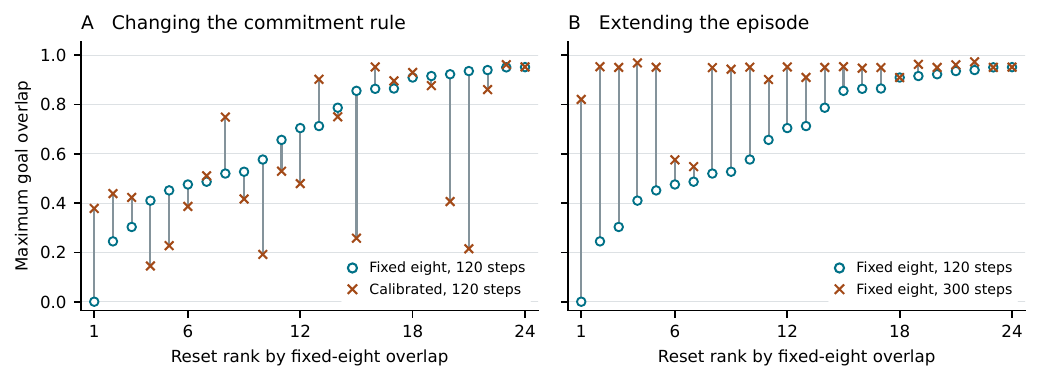}
\caption{Frozen Diffusion Policy on the same 24 test resets. Open circles in both panels denote fixed eight-step commitment with a 120-action budget. Each vertical segment joins observations from the same reset. Changing the commitment rule improved overlap on some resets and reduced it on others. Extending fixed-eight execution to 300 actions allowed the same controller more time to reach the goal.}
\label{fig:dphorizon}\end{figure*}

\begin{table}[!t]\centering\caption{Public image-conditioned policies, 24 shared test resets. Peak is the episode-mean maximum goal overlap. Calls are per 100 executed actions. Exc.\ counts episodes with an authorized 32-pixel exceedance.}\label{tab:visual}
\footnotesize\begin{tabular}{@{}llrrrr@{}}\toprule
Model & Commitment & Success & Peak & Calls & Exc.\\\midrule
SmolVLA & Fixed eight & 0/24 & 0.115 & 12.5 & 18\\
 & Demo threshold & 0/24 & 0.121 & 24.1 & 3\\
 & Policy threshold & 0/24 & 0.092 & 39.6 & 0\\
Diffusion & Fixed eight & 2/24 & 0.665 & 12.5 & 21\\
 & Policy threshold & 3/24 & 0.577 & 39.7 & 0\\
\bottomrule\end{tabular}\end{table}
SmolVLA completed none of the 24 tasks under any of the three modes. Its fixed-eight peak overlap averaged 0.115. The frequency of geometric exceedances can consequently improve even for a controller that seldom approaches the task goal. Fig.~\ref{fig:smol} shows uncertainty in the paired contrasts. With Diffusion Policy, fixed-eight execution reached the target in 2 of 24 episodes and the calibrated rule in 3. Their numbers of violating episodes were 21 and 0, at 12.5 and 39.7 calls per 100 actions. The calibration samples and success counts are too small for a task-success noninferiority claim. The published checkpoints have different architectures, training budgets and action samplers; these data compare execution rules within each checkpoint.
Across the 24 matched resets at 120 steps, the two fixed-eight successes also succeeded under the calibrated rule. One additional reset succeeded only under calibration. In a post hoc budget check, extending the fixed-eight Diffusion Policy runs on the same resets to 300 actions increased mean peak overlap from 0.665 to 0.910 and yielded 16 successes. Fig.~\ref{fig:dphorizon} shows the paired changes in overlap. The longer budget was tested only for the fixed-eight controller.

\section{How Many Calibration Episodes Buy a Useful Prefix?}
Two controllers can meet the same tolerance by very different means. One predicts the motion accurately over most of the chunk. The other stops immediately whenever it sees an unfamiliar configuration. Both can have zero authorized violations on a test set. They have different costs when the robot must actually move. The finite calibration rank puts a precise limit on how much evidence is available to distinguish these cases.

\subsection{Rank resolution before running the policy}
The largest finite calibration score can be used only when $\lceil(m+1)(1-\alpha)\rceil\leq m$, equivalently $\alpha\geq1/(m+1)$. Thus a distribution-free target of 10\% admits a finite rank starting at nine calibration episodes. A 5\% target requires at least 19. If the corpus contains thousands of decision windows but only eight independent episodes, subdividing it cannot increase $m$ for the complete-episode event in Proposition~1. At the minimum calibration size the finite threshold is the largest observed episode score. One atypical push can then shorten recommendations across the entire deployment sample.

There are two distinct sources of variation. Conditional on the realized calibration set, the test performance of its rank threshold can deviate from the nominal level. Across possible calibration sets, the threshold itself changes. For continuous independent episode scores with common distribution function $F$, the transformed order statistic $F(S_{(k)})$ has the $\mathrm{Beta}(k,m+1-k)$ distribution. This follows by counting how many of the $m$ uniform transforms fall below a given value. Its mean $k/(m+1)$ recovers the familiar marginal rank statement. Its variance is $k(m+1-k)/[(m+1)^2(m+2)]$. This identity describes score coverage before selecting a prefix. The realized probability of exceeding a positional tolerance among released actions can be smaller. A test-set fraction of violating episodes therefore does not estimate the rank threshold's quantile index.

We examined threshold variation with the original predictor and scale frozen in each corpus. From the 60 primary calibration episodes, we drew 160 random permutations. The first 8, 12, 20, 40, or 60 episodes of a permutation formed nested calibration sets. We retained the same 60 test episodes, seven tolerance settings in the primary analysis, and the graph-dependent scale. Here we report the previously tabulated operating point of 32 pixels for Push-T and 2 mm for the physical rectangle. The only choices varied in this analysis were calibration size and nominal risk (5\%, 10\%, or 20\%). Identical test resets and overlapping calibration subsets make the comparisons paired. The 160 draws describe sensitivity to which records were available in these finite corpora; they do not create 160 independent deployment studies.

\begin{figure*}[!t]\centering\includegraphics[width=.94\textwidth]{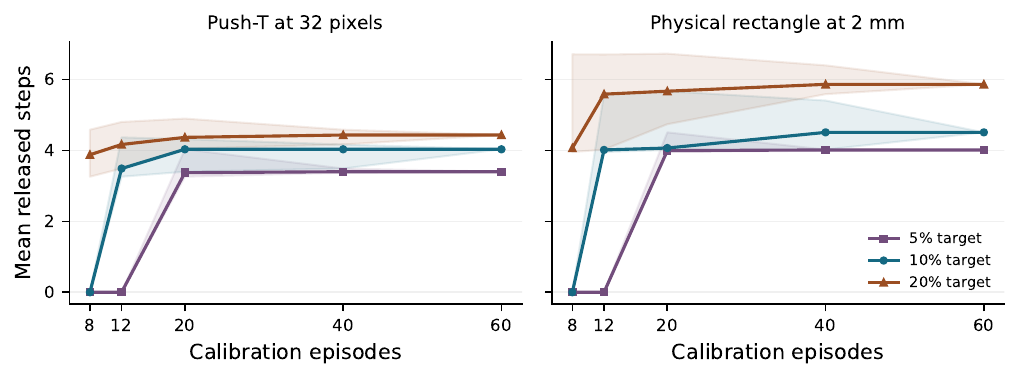}
\caption{Calibration-size sensitivity with a frozen forecaster. Each point gives the median test-set mean prefix over 160 nested subsamples of the calibration pool. Shading spans the 10th to 90th percentiles across subsamples. For 60 episodes, every permutation contains the same observations. A zero at small calibration sizes follows exactly from the infinite-threshold convention.}\label{fig:calbudget}\end{figure*}

\begin{table}[!t]\centering\caption{Sensitivity at a 10\% target. Prefix ranges are the 10th and 90th percentiles over calibration subsets. The last column gives the corresponding range of counts of violating test episodes out of 60.}\label{tab:calbudget}
\footnotesize\begin{tabular}{@{}lrrcr@{}}\toprule
Corpus & $m$ & Median prefix & Prefix range & Violations\\\midrule
Push-T & 12 & 3.49 & 3.26--4.37 & 1--4\\
       & 20 & 4.03 & 3.40--4.32 & 1--4\\
       & 40 & 4.03 & 3.49--4.17 & 1--4\\
       & 60 & 4.03 & 4.03--4.03 & 4--4\\
Physical & 12 & 4.02 & 3.96--5.59 & 1--10\\
         & 20 & 4.07 & 4.02--5.67 & 1--10\\
         & 40 & 4.51 & 4.03--5.41 & 1--9\\
         & 60 & 4.51 & 4.51--4.51 & 5--5\\\bottomrule
\end{tabular}\end{table}

\subsection{What changes when the rank becomes finite}
At eight calibration episodes and a 10\% target, every branch in both replays receives an infinite bound. All prefixes have length zero. At 12 episodes the rank becomes finite but still selects the maximum of the 12 scores. On Push-T the median mean prefix among those subsets was 3.49 actions. Its central 80\% range was 3.26--4.37. Raising $m$ to 40 narrowed this range to 3.49--4.17. Both experiments reuse the same test labels, so the narrowing reflects variation in the estimated threshold alone. With all 60 calibration episodes, the 10\% rank is 55 and the primary mean of 4.03 is recovered exactly.

The physical data exhibit a different sensitivity at the same nominal risk. With 20 calibration episodes, central subsets produced between 4.02 and 5.67 steps on average. On the same test pushes, the counts of violating episodes ranged from one to ten within the central 80\% of subset results. At $m=40$ that range was still one to nine. Discrete steps and gaps between high episode scores make the mapping from a quantile to the operating point uneven. Doubling the number of calibration trajectories cannot be assumed to double usable commitment, and it need not improve every observed violation count monotonically. A pooled table that retained only the median would conceal this dependence on which motions had been calibrated.

Changing the target also exposes a practical difference between the tasks. With all 60 physical calibration episodes, moving from 10\% to 20\% increased mean commitment from 4.51 to 5.87 steps and doubled the number of observed violating episodes from five to ten. On Push-T the same change gave 4.44 instead of 4.03 steps, with four observed violating episodes at both settings. Equality of those counts on one test set does not establish equality of population risks. It shows that the extra authorized steps did not introduce a previously unaffected violating episode in this particular set: the authorized prefix nests as the threshold decreases.

\subsection{Reading a decision before and after execution}
\begin{figure*}[!t]\centering\includegraphics[width=.94\textwidth]{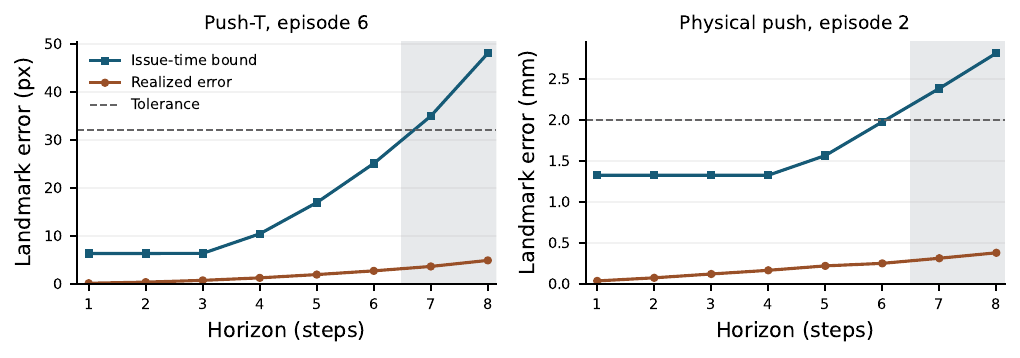}
\caption{Two logged decisions under the full calibration set. The shaded region begins at the first unauthorized step. Bounds and realized errors share physical units within a panel. The y-axes differ because the first task measures image-plane landmarks in scene pixels, whereas the second reconstructs metric probe locations.}\label{fig:window}\end{figure*}

Figure~\ref{fig:window} shows the first partially authorized test window in chronological order within ascending episode IDs for each corpus. At Push-T episode 6, anchor 1, the calibrated bounds at the sixth and seventh horizons were 25.09 and 34.96 pixels. With a 32-pixel tolerance, the first six actions qualified. The realized errors at these same horizons were 2.69 and 3.61 pixels. A reader who sees the outcome can identify a seventh action that would also have met the tolerance along the recorded trajectory. The controller at issue time had access only to the bounds and the proposed commands. Retrofitting the stopping point to the realized error would change both the decision information and the reported event.

For physical episode 2, anchor 1, the sixth-step bound was 1.975 mm, only 0.025 mm below tolerance. The seventh-step bound reached 2.383 mm. The corresponding realized errors were 0.249 and 0.311 mm. These small errors are consistent with a straight push whose motion was easier to forecast than the cases determining the upper calibration rank. The example also specifies the precision needed for reproducibility: comparisons are made before rounding bounds for display. At a configured tolerance near a bound, a controller comparing rounded millimeters could release a different prefix from the one evaluated here.

\subsection{From prefix length to a control budget}
A replayed mean prefix is an episode-weighted mean over prespecified windows. An online policy makes its next query when its released actions run out, when it executes a fallback, or when the task terminates. The reciprocal of the mean replay prefix is therefore not an estimator of online query rate. Define $N_i$ as executed actions, $Q_i$ as policy queries, and $F_i$ as fallback actions on online episode $i$. We compute calls per hundred actions as $100\sum_i Q_i/\sum_i N_i$. This counts incurred work, whereas averaging $Q_i/N_i$ over resets would weight a short successful episode as heavily as one that used the entire budget.

For the retrieval policy, demonstration calibration produced 2,694 queries for 9,299 actions; policy calibration required 5,042 queries for 9,143 actions. The latter total contains 31 one-step fallbacks. For SmolVLA the two calibrated modes executed the same 2,880 actions each, with 695 versus 1,141 queries and two versus 17 fallbacks. In those trajectories, including fallback actions in the violation check happened to leave the episode counts unchanged. The logical distinction remains important. The certificate is issued for a selected positive prefix. An implementation operating through an empty prefix takes a separate action for which that certificate offers no bound.

The cost of a query also depends on the controller. The visual diffusion test executes 100 denoising steps per proposal; an extra call there consumes different resources from a nearest-neighbor lookup in a fixed demonstration collection. Counting calls gives an architecture-independent description of when the rule requests information. Converting the counts to wall-clock deadlines would additionally require measured sensor and proposal latency, batching policy and the ability to overlap planning with execution. The closed-loop data support separate comparisons within each controller. The offline rank experiment and the hardware replay establish how the calibration evidence behaves under their own data collections.

\section{Conditions for Use on a Robot}
A human demonstrator may choose subsequent commands in response to observations seen after the start of a logged chunk. A deployed VLA cannot know those observations at issue time. The study therefore estimates the relation between a given chunk and its recorded consequence; it does not identify the causal effect of committing that chunk under an alternative policy. In closed-loop Push-T, the same reset can visit different states under different commitments. This explains why an episode maximum calibrated on demonstrations requires a new audit at deployment.

Transfer between tasks requires similar care. LIBERO tests multi-task transfer \cite{libero}, CALVIN emphasizes long sequences of language-conditioned behavior \cite{calvin}, and RLBench and Meta-World offer varied manipulation configurations \cite{rlbench,metaworld}. Observational datasets such as BridgeData V2 and studies of human demonstration processing document the variety of available collection regimes \cite{bridge,robomimic}. Calibrating on one mixture and testing on another can change episode length and the frequency of rare contacts. A deployment evaluation should preserve task and operator identifiers so these shifts can be measured. The two replay studies cover two planar object shapes under separate collection protocols; both use supplied pose information and a fixed anchor grid.

\section{Conclusion}
The largest practical difference in this replay came from the choice of calibration unit. Aggregating residuals across a complete trajectory shortened the admissible prefix and reduced the number of episodes with a positional violation. Within that rule, geometry-dependent scales recovered some of the lost horizon on Push-T. The flat scale admitted longer prefixes on the physical straight pushes. With the frozen retrieval policy, recalibration on separate simulator resets reduced the observed count of violating episodes from 15 to two and increased the query rate from 29.0 to 55.1 per 100 actions. On the frozen diffusion controller, the calibrated rule eliminated the 21 observed geometric exceedances at a 120-action budget while increasing policy calls. Extending fixed-eight execution alone from 120 to 300 actions raised completion from two to 16 of 24 resets. Deployment claims therefore need a task budget as well as a forecast tolerance.

\section*{Code and Data Availability}
The \href{https://github.com/LudwigWangMu/episode-calibrated-action-prefixes/tree/8615bb18927b8d9e30e2e9209c73ed652b489985}{\nolinkurl{episode-calibrated-action-prefixes}} repository contains the split assignments, per-episode outcomes, and fixed revisions of both visual policies. Run \texttt{python prepare\_inputs.py} in that directory to reconstruct the Push-T and MCube inputs, then \texttt{python reproduce.py} for the CPU experiments. The README gives the separate GPU commands. From the repository root, \texttt{python verify.py} checks the archived policy results without downloading weights. The archived version is \href{https://github.com/LudwigWangMu/episode-calibrated-action-prefixes/releases/tag/v1.0.1}{release v1.0.1}.

\section*{Acknowledgment}

\IEEEtriggeratref{38}
\bibliographystyle{IEEEtran}
\bibliography{references}
\end{document}